\documentclass{midl} %

\usepackage{adjustbox}
\usepackage{tikz}
\renewcommand{\theta}{\vartheta}
\renewcommand{\epsilon}{\varepsilon}

\usepackage{placeins} %
\usepackage{float} %
\usepackage{bm}
\usepackage{multirow}

\usepackage{mwe} %
\jmlryear{2023}
\jmlrworkshop{Full Paper -- MIDL 2023}

\usepackage{xcolor}

\renewcommand{\theta}{\vartheta}

\midlauthor{\Name{Florentin Bieder\nametag{$^{1}$}} \Email{florentin.bieder@unibas.ch}\\
\Name{Julia Wolleb\nametag{$^{1}$}} \Email{julia.wolleb@unibas.ch}\\
\Name{Alicia Durrer\nametag{$^{1}$}} \Email{alicia.durrer@unibas.ch}\\
\Name{Robin Sandk\"uhler\nametag{$^{1}$}} \Email{robin.sandkuehler@unibas.ch}\\
\Name{Philippe C. Cattin\nametag{$^{1}$}} \Email{philippe.cattin@unibas.ch}\\
\addr $^{1}$ Department of Biomedical Engineering, University of Basel \\
}

\title[PatchDDM]{
Memory-Efficient 3D Denoising Diffusion Models for Medical Image Processing}
\begin{document}

\maketitle

\begin{abstract}
Denoising diffusion models have recently achieved state-of-the-art performance 
in many image-generation tasks. They do, however, require a large amount of 
computational resources. This
limits their application to medical tasks, where we often deal 
with large 3D volumes, like high-resolution three-dimensional data. In this work, we 
present a number of different ways to reduce the resource consumption for 3D 
diffusion models and apply them to a dataset of 3D 
images.
The main contribution of this paper is the memory-efficient patch-based diffusion model  \textit{PatchDDM}, which can be applied to the total volume 
during inference while the training is performed only on patches. 
While the proposed diffusion model can be applied to any image generation tasks, 
we evaluate the method on the tumor segmentation task of the BraTS2020 
dataset and demonstrate that we can generate meaningful 
three-dimensional segmentations.
\end{abstract}


\section{Introduction}

Denoising diffusion models \cite{HoJainDDPM2020, nicholDhariwal21a} have 
lately shown an impressive performance in image generation and experienced 
increasing popularity in medical image analysis 
\cite{kazerouni2022diffusion}. However, the processing of large three-dimensional (3D) 
volumes, which often is required in medical applications, is still a challenge. 
Limitations related to the computational resources only allow the processing of small 
3D volumes, which impedes the processing of high-resolution magnetic resonance 
(MR) or computer tomography (CT) scans.

\begin{figure}[htbp]
\floatconts
  {fig:overview}
  {\caption{Overview of our proposed method \textit{PatchDDM}. The diffusion model is optimized in memory efficiency and speed by training only on coordinate-encoded patches. The input consists of noised $x_t$, the volumes $b$ that are to be segmented and which are provided as a condition for the segmentation, as well as a coordinate encoding $CE$ for the patches. During sampling, the whole 3D volume can be processed at once.}}
  {\includegraphics[width=0.77\linewidth]{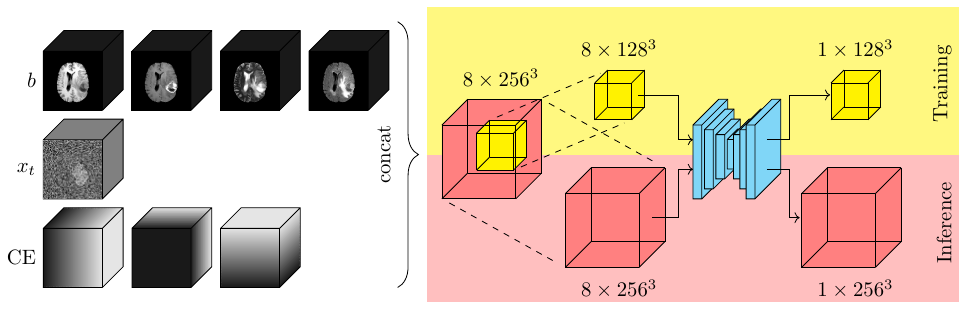}}
\end{figure}

\paragraph{Contribution}
In this work, we introduce architectural changes to the state-of-the-art 
diffusion model implementation \cite{nicholDhariwal21a}, enabling to train 
on large 3D volumes with commonly available GPUs. 
 We adapt the U-Net-like architecture to improve the speed and memory efficiency. 
 Furthermore, we propose a novel method illustrated in Figure \ref{fig:overview}.
 With this method, the diffusion model is trained only on coordinate-encoded patches 
 of the input volume, which reduces the memory consumption and speeds 
 up the training process. During sampling, the proposed method allows 
 processing of large volumes in their full resolution without needing to split them up into patches.
To evaluate our method, we perform
diffusion model based image segmentation \cite{wolleb2022diffusion}
that has previously been proposed for 2D segmentation
on the BraTS2020 dataset \cite{menze2014multimodal, bakas2017advancing, bakas2018identifying}.

\paragraph{Related Work}
Denoising diffusion models have seen a quick adoption in research,
replacing the more traditional generative models in many tasks such as 
such as unconditional and conditional image generation \cite{HoJainDDPM2020, song2021denoisingDDIM, nicholDhariwal21a},
text-to-image translation \cite{nichol2021glide, saharia2022imagen, ramesh2021dalle, Kim2022clip}
and inpainting \cite{ramesh2021dalle, nichol2021glide}.
Diffusion models have also been used for various applications in the medical field, for instance, for anomaly detection \cite{wolleb2022anomaly}, 
synthetic image generation \cite{dorjsembe2022medicalimagesynthesis, peng2022generating} 
and segmentation \cite{guo2022accelerating, medsegdiff2022, wolleb2022diffusion}. 
Medical images, however, often are 3D volumes, such as MR- or CT-scans. These volumes create challenges regarding memory consumption of processing methods.
Consequently, many of the current methods are limited to two-dimensional (2D) slices only \cite{medsegdiff2022, wolleb2022diffusion, guo2022accelerating}
or to 3D volumes restricted to a limited resolution of at most $128 \times 128 \times 128$ \cite{khader2022medical, peng2022generating, dorjsembe2022medicalimagesynthesis}.
To the best of our knowledge, we are the first to tackle the challenge of applying denoising diffusion models to large 3D volumes.

\section{Method}

We explore how denoising diffusion implicit models (DDIMs) presented in 
Section \ref{sec:ddim} can be improved regarding memory efficiency 
and time consumption. The required architectural changes are presented 
in Section \ref{sec:architecture}. We present the training and sampling 
scheme of our method \textit{PatchDDM} in Section \ref{sec:patches}. 
For evaluation, we use the segmentation approach using denoising diffusion 
models presented in Section \ref{sec:segmentation}.

\subsection{Denoising Diffusion Models}\label{sec:ddim}
In the following, we will use the notation introduced by \cite{HoJainDDPM2020}.
Denoising diffusion models
rely on an iterative noising and denoising process.
The forward noising process $q$
is given by 
\begin{equation}
    q(x_t \mid x_{t-1}) = \mathcal N(x_t; \sqrt{1-\beta_t}x_{t-1}, \beta_t I)  \label{eq:forward_noising}\\
\end{equation}
where $\beta_t$ is a predefined sequence of variances. 
We can directly compute $x_t$ from a given $x_0$ with 
\begin{equation}
\label{eq:forward_relation}
    q(x_t \mid x_0) := \mathcal N(x_t; \sqrt{\overline{\alpha_t}} x_0, (1-\overline{\alpha_t}) I) 
\end{equation}
with $\alpha_t := 1 - \beta_t$ and $\overline{\alpha_t} := \prod_{s=1}^t \alpha_s$. 
This corresponds to degrading the input image by adding Gaussian noise. 
For image generation tasks we are interested in the reverse process $p_\theta$. 

\begin{equation}
    p_{\theta, t}(x_{t-1} \mid x_t) = \mathcal N(x_{t-1}; \mu_{\theta, t} (x_t), \Sigma_{\theta, t} (x_t))
\end{equation}
Both $\mu_{\theta, t}$ and $\Sigma_{\theta, t}$ can be estimated by a U-Net-based network $\epsilon_{\theta, t}$ with parameters $\theta$. 
The loss used to train the network $\epsilon_{\theta, t}$ can be written as
\begin{equation}\label{eq:mse}
{\Vert \epsilon-\epsilon_{\theta, t}(x_t,t)\Vert}^{2} 
= {\Vert\epsilon-\epsilon_{\theta, t}(\sqrt{\overline{\alpha_t}}x_0+\sqrt{1-\overline{\alpha_t}}\epsilon,t)\Vert }^{2}, 
\quad \text{with } \epsilon \sim \mathcal{N}(0,I).
\end{equation}
Using the
DDIM \cite{song2021denoisingDDIM} sampling scheme, we can define
\begin{equation} %
\label{eq:backward_denoising}
    x_{t-1} = \sqrt{\overline{\alpha_{t-1}}}
    \left( \frac{x_t - \sqrt{1-\overline{\alpha_t}} \epsilon_{\theta, t} (x_t)}{\sqrt{\overline{\alpha_t}}} \right)
    + \sqrt{1-\overline{\alpha_{t-1}}} \epsilon_{\theta, t} (x_t) ,
\end{equation}
where $\epsilon_{\theta}(x_t)$ is the output of the network.
This sampling scheme has the advantage that the denoising process 
is deterministic and we do not need to sample
random vectors in every step. Thus, the only source of 
stochasticity during inference originates from the
random initial sample $x_T$ which is sampled from $\mathcal N(0, I)$.
During inference, a sequence of images $x_i$ for $i = T, T-1, \ldots, 0$ 
of decreasing noise level is being generated, the initial
$x_T$ is sampled from a standard normal distribution $\mathcal N(0, I)$.

\subsection{Architecture} \label{sec:architecture}

We adapt the 2D-U-Net-based network architecture 
proposed by \cite{HoJainDDPM2020, nicholDhariwal21a} and
used by \cite{nichol2021glide, coldDiffusion22, medsegdiff2022, song2021denoisingDDIM} for the application on 3D data.
The previously proposed architecture
features two or three residual convolutional blocks at each down- and upsampling step.
Furthermore, it uses attention blocks at multiple resolutions as well as in the bottleneck. 
 \cite{pallette22} determined that adding global self attention 
 can slightly improve the quality 
of the generated images as compared to an increase in convolutional blocks.
For 3D data, the attention blocks use disproportionally more memory,
which made it infeasible to use on current hardware, which is why
we removed them completely.
\begin{figure}[htbp]
\floatconts
  {fig:architecture}
  {\caption{The architecture of the U-Net-like network with averaging skip connections. In the original network as well as in the U-Net the $\bigoplus$ operator is a concatenation $x = (x_s, x_u)$, in our case it is an averaging operator $x = (x_s + x_u)/2$.}
  }
  {
  \includegraphics[width=0.3\linewidth]{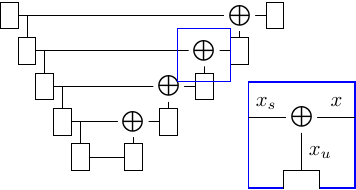}
  }
\end{figure}
The second fundamental change we implemented was the use of additive skip connections,
as shown in Figure \ref{fig:architecture}.
In the previous architecture as well as in the original U-Net implementation \cite{ronneberger2015u}, the skip connection uses concatenation to combine $x_s$
from the encoder with the upsampled tensors $x_u$ from the lower resolution path of the decoder.
This implies that the decoder requires significantly more resources than the encoder, especially
at the highest resolution levels. To alleviate this issue, we propose to 
average them as $x = \frac12 (x_s + x_u)$.
Unlike in ResNet \cite{resnet2016}, where the skip connections 
are added, we found the averaging to be crucial 
for avoiding numerical issues like exploding gradients.
Intuitively this can be justified by considering $x_s$ and $x_u$ as
random variables with $x_u, x_s \text{ iid. } \mathcal N(0, \sigma^2)$. 
Therefore, $x_u + x_s \sim \mathcal N(0, 2\sigma^2)$,
that is, with each concatenation the variance doubles, while
averaging preserves the variance. 

The savings in memory from replacing the concatenation with the averaging allow us to
increase the network width, i.e. the number of channels within the whole network, by a
factor of 1.61, while preserving the total memory usage. %
Furthermore, the resulting network architecture allows for training on varying input sizes.
This property is crucial or our proposed patch-based method.
For all of our experiments, we use the same network configuration.

\subsection{Patch-based Approach with Coordinate-encoding}\label{sec:patches} 

To benefit from the lower requirements of computational resources but 
still to operate on the original
resolution, we propose a novel patch-based training method named 
\emph{PatchDDM} that trains on randomly sampled patches of the input
but can afterwards be applied to the full resolution volume during inference.
This means for the training we can benefit from using smaller inputs,
which means we need less computations per iteration as well as less memory.
For the inference, however, we can pass the entire input volume at once 
\emph{without} having to sample patches and reassemble them.
This means no boundary artifacts do to the separate padding of the patches
within the CNN are introduces, and also the stitching artifacts that
that can appear in traditional patch-based approaches are eliminated.

To add information about the position of the patch, we condition the network on
the position of the sampled patch. We implemented this by concatenating a grid of
Cartesian coordinates to the input. Each coordinate is represented by one channel as
a linear gradient ranging from -1 to 1.
This is similar to the method proposed in \cite{coordConv}.
They propose to add the coordinates as additional channels for before all convolutions.

In our case, we append the coordinates to the whole input just like in \cite{coordConv}, but then sample a patch,
where the coordinates serve as a position encoding for the sampled patch.
An overview of this coordinate encoding is given in Figure \ref{fig:overview}.
For the BraTS2020 data, the subject is centered within the volume.
We use a patch sampling strategy, assigning a higher probability to the center of the volume,
as shown in Figure \ref{fig:patchsampling} in the Appendix \ref{sec:samplingdistribution}.%

\paragraph{Baseline methods}\label{sec:fulres}
For our ablation study, we use two baselines with the same network  %
as our proposed approach, but without patch-based training.
Furthermore we also performed an experiment with our patch based approach
but without the proposed coordinate encoding. The training did not converge
and did not produce any usable result. Therefore, we will not report any
metrics from this experiment. The two baseline methods are the following:

\begin{itemize}
\item Training on full resolution (\emph{FullRes}):
We implemented a distributed version of the proposed architecture that splits the task to two GPUs if necessary. This allows for training directly on full resolution ($256^3$) data, given that the expensive specialized GPU hardware is available.

\item Training on half resolution (\emph{HalfRes}):
A straightforward way to reduce the requirements in 
terms of computational resources is training
the model on downsampled data. In our experiments, we 
downsampled the input image before passing it to the network,
but then upsampled the output of the network again to evaluate 
the performance on the full size.
For three spatial dimensions (i.e. 3D) this means that reducing the 
input size from $256^3$ to $128^3$ results in a reduction 
of a factor of $8$ in terms of memory and computation time, 
allowing this model to be run on widely available GPUs.

\end{itemize}

\subsection{Denoising Diffusion Models with Ensembling for Segmentation}\label{sec:segmentation}
In order to generate the segmentation of an input image $b$,
we need to condition the generation of the segmentation mask $x_0$ on that given image $b$.
 We will follow the method
proposed by \cite{wolleb2022diffusion}, where the input images $b$ are being concatenated
to every $x_t$ as a condition. 
It was shown that ensembling several predicted segmentation masks per 
input image increases the segmentation performance \cite{segdiff2021, wolleb2022diffusion}. 
An overview of this segmentation approach is given in  Figure \ref{fig:segmentationprocess}. 
An advantage of the denoising diffusion based segmentation approach
is the implicit ensembling we get when using different samples $x_T$ from the
noise distribution  $\mathcal N(0, I)$, which can be used to increase
the performance and estimate the uncertainty.
To evaluate the performance of our proposed method \emph{PatchDDM} described in Section \ref{sec:patches} 
and the two baseline methods \emph{FullRes} 
and \emph{HalfRes}, we apply
our method to a segmentation task as proposed in \cite{wolleb2022diffusion}.
Therefore, we train our diffusion model to generate semantic segmentation masks. 

\begin{figure}[htbp]
\floatconts
  {fig:segmentationprocess}
  {\caption{The ground truth segmentation $x_0$ is degraded by the noising process $q$. 
  We train a network to perform the denoising process $p_\theta$, that is, given some noised image $x_t$,
  we train it to denoise it with the MR-sequences $b$ as a condition.}}
  {\includegraphics[width=0.65\linewidth]{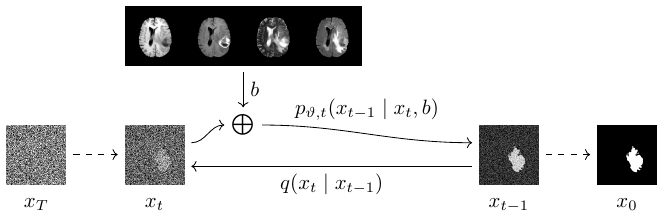}}
\end{figure}

\section{Experiments}\label{sec:experiments}

\paragraph{Dataset}
For our experiments, we used the BraTS2020 dataset \cite{menze2014multimodal, bakas2017advancing,  bakas2018identifying}.
It contains 369 head MR-scans, each including four sequences (T1, T1ce, T2, FLAIR) 
with a resolution 
of $1 \times 1 \times 1 \text{ mm}^3$, resulting in a total scan size of $240\times 240 \times 155$, which we padded to a size of $256\times 256 \times 256$.
The background voxels were set to zero and the range between the first and 99th percentile was normalized
to $[0,1]$.
We used an $80\%/10\%/10\%$ split for training, validation and testing.
The label masks consist of three classes, namely the Gadolinium-enhancing tumor,  %
the peritumoral edema, %
and the necrotic and non-enhancing tumor core. %
For the binary segmentation experiments, all three classes were merged into one.

\paragraph{Training Details}

We performed our experiments on NVIDIA A100 GPUs with $40GB$ of memory each. 
To directly train on the full resolution
$256^3$ images, we distribute the model over 2 GPUs. The methods \emph{HalfRes} and \emph{PatchDDM} were trained on one GPU only.
The optimizer we used was AdamW \cite{adamw} with the default parameters.
We chose the learning rate $\operatorname{lr} = 10^{-5}$ by optimizing the average 
Dice coefficient on the validation
set after 150k optimization steps over a range of values 
between $10^{-6}$ and $10^{-3}$.
We trained the models for the same amount of time for all experiments (420h). For
the evaluation, we selected the best-performing models based on the average Dice score
on the validation set based on a single evaluation, i.e., without ensembling.
For the denoising process, we set the number of steps to $T = 1000$ and
use the affine variance schedule proposed in \cite{HoJainDDPM2020} 
with $\beta_1 = 0.02, \beta_T = 10^{-4}$. 
\paragraph{Accelerated Sampling}
By default we need $T = 1000$ denoising steps for the inference.
As shown in \cite{song2021denoisingDDIM}, we can interpret the the DDIM denoising step \eqref{eq:backward_denoising}
 as the Euler discretization of an ordinary differential equation (ODE).
This insight motivates the use of larger step sizes with respect to $t$ during inference,
which allows for accelerated sampling. 
The drawback is that the output quality deteriorates
with fewer samples. We investigate how we can trade off
fewer sampling steps (larger step sizes) and ensembling (more samples).

\section{Results}
In the following, we will assess the performance of our proposed model and compare it
to the two baseline approaches.
For each model, we computed the average Dice score on the validation set and used this
to choose the best-performing checkpoint.
We provide some qualitative outputs in Figure \ref{fig:qualitativeoutputs} in the Appendix \ref{sec:qualitativeresults}.
To assess the training progress, we display the Dice score as well as the HD95 
(Hausdorff distance, 95th percentile) of \emph{PatchDDM} over the course of the training 
in Figure \ref{fig:patch_training} in the Appendix \ref{sec:trainingprogress}. 
The metrics of best-performing checkpoint with respect to the
Dice score when using a single evaluation (no ensembling) is reported in 
Table \ref{tab:singlescores} in Appendix \ref{sec:singleevaluationscores} along with the score of the state of the art nnU-Net \cite{isensee2021nnu}.

\subsection{Segmentation Ensembling}
To evaluate the impact of ensembling, we 
compute the Dice- and HD95-score of the three methods (\emph{PatchDDM}, \emph{FullRes}, \emph{HalfRes})
with resepect to ensemble size, see Figure \ref{fig:ensemblingperformance}. Both scores
significantly improve using ensembles
for our proposed \emph{PatchDDM} and the \emph{FullRes} method.
 In Table \ref{tab:ensemblingscores}
in the Appendix \ref{sec:ensemblingscores} the metrics for different ensemble sizes are provided.
The curves show that ensembling can further improve
the performance and get very close to the best performing ensembles
with an ensemble size of as small as five to nine.

\begin{figure}[htbp]
\floatconts
  {fig:ensemblingperformance}
  {\caption{The evaluation metrics on the test set as a function of the ensemble size.}}
  {\includegraphics[width=0.7\linewidth, clip, trim=3mm 2mm 3mm 1mm]{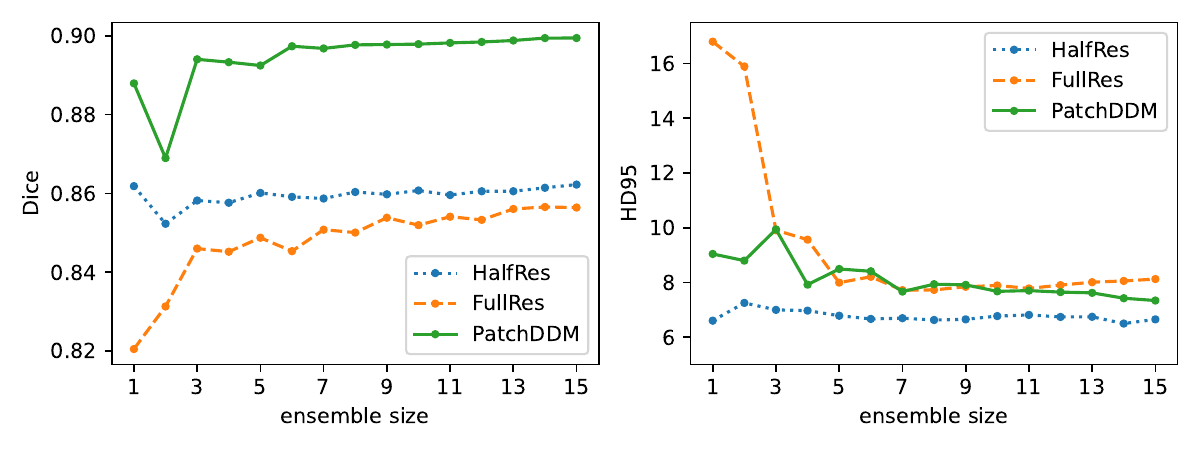}}
\end{figure}

\subsection{Computational Resources \& Time Requirements}
We report the memory consumption and the time required for one model evaluation for all comparing methods. As displayed in Table \ref{tab:resources},
the training of \emph{FullRes} needs close to 80GB of memory.
This requires at the time of writing still highly expensive hardware.
The other baseline \emph{HalfRes} as well as our proposed \emph{PatchDDM} method
both need less than 12GB for training and can therefore be trained
on much cheaper and widely available hardware. The reduced resolution
also results in a reduction in the number of computations, and
therefore a larger number of optimization steps that can be performed
in a given time interval.
A drawback of our proposed method is the increased memory consumption
and reduced speed during inference, both of which are comparable to the
\emph{FullRes} model.

 \begin{table}[htbp]
	\floatconts
	{tab:resources}%
	{\caption{Memory consumption in GB and time in seconds for one network evaluation. The memory requirements for the distributed run also includes a small amount of overhead, 
     as some arrays are duplicated on both GPUs.}}%
	{\begin{tabular}{l|ll|ll}
               & \bfseries Memory &  & \bfseries Time &  \\
            \bfseries Method & \bfseries Training  & \bfseries Inference & \bfseries Training & \bfseries Inference \\ \hline
			\emph{FullRes}                &  78.5  &    25.7  &  2.12  &  1.01 \\
            \emph{HalfRes}              &     10.5 &    \hphantom{0}4.90  &  0.351 &  0.124 \\
            \emph{PatchDDM}           &     10.6 &    24.0  &  0.340  &  1.02 \\
	\end{tabular}}
\end{table}

\subsection{Ensembling and Accelerated Sampling}
Figure \ref{fig:acceleratedsampling} shows the trade-off between
the ensemble size and the number of sampling steps. With as little
as 20 sampling steps (i.e. a step size of 50), the performance
is already close to the results obtained with $T=1000$ steps, implying a speedup of
a factor of $50$. But even with fewer step sizes, we can trade
the number of steps for a greater ensemble size to achieve a similar performance.
Consequently, for a fixed budget of network evaluations (i.e. steps), we can profit
from using ensembling with accelerated sampling.

\begin{figure}[htbp]
\floatconts
  {fig:acceleratedsampling}
  {\caption{The average Dice score and HD95 metric on the test set as a function of the number of 
  sampling steps and the ensemble size. The white sections indicate that we did not evaluate that combination.}}
  {\includegraphics[width=0.7\linewidth]{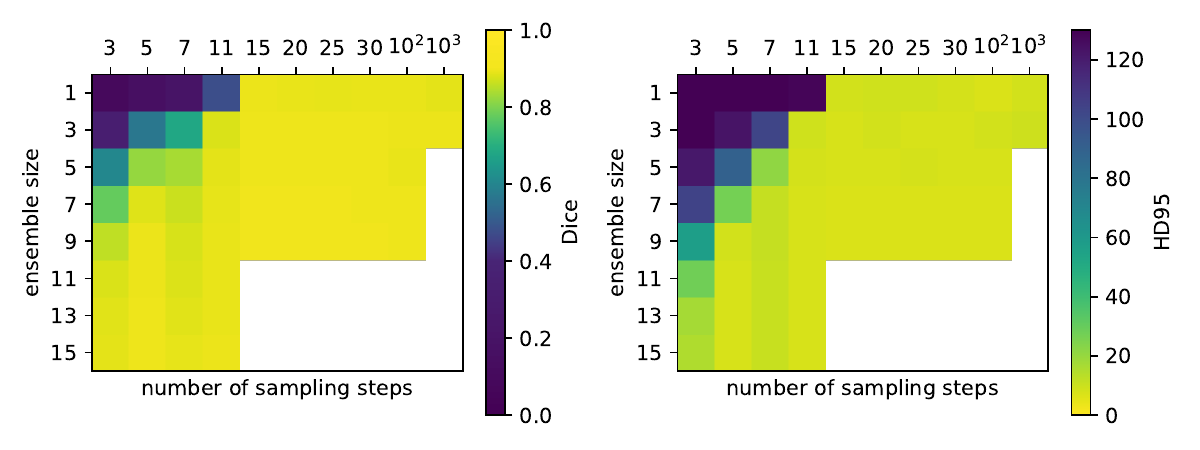}}
\end{figure}

\section{Discussion}
We propose \emph{PatchDDM}, a novel patch-based diffusion model architecture that allows the training of diffusion
models on high-resolution 3D datasets. This enables denoising diffusion
models to be used for image analysis and -processing tasks 
in medicine on commonly available hardware.
We could demonstrate the effectiveness by
applying it to a recently developed segmentation framework
for medical images.
In the future, we would like to investigate the performance
of our proposed approach for tasks involving image generation. Furthermore, we will investigate 
the role of the patch
size used and whether it can be made smaller for processing
even higher resolution volumes.
In order to preserve high quality, \cite{karras2022elucidating} proposed
using higher-order ODE solvers, like the Heun, method when choosing
larger step sizes. This might further reduce the number of iterations needed.
Finally, it would be interesting to investigate an extension
of this segmentation framework that includes
multiple classes.

\midlacknowledgments{
We are grateful for the support of the Novartis FreeNovation
initiative and the Uniscientia Foundation (project \#147-2018).
We would also like to thank the NVIDIA Corporation for donating a GPU that was used for our experiments.
}
\newpage
\bibliography{bieder23}

\newpage
\appendix

\FloatBarrier
\section{Sampling Distribution}\label{sec:samplingdistribution}
\FloatBarrier
\begin{figure}[htbp]
\floatconts
  {fig:patchsampling}
  {\caption{The sampling distribution was chosen empirically to favor the central patches.
  The distribution  $p$ is defined over the normalized coordinates of admissible patches (normalized to $[-1, 1]$)
  and can be interpreted as the probability density function the sum $X+Y$ 
  of two random variables $X \sim U[-1/3, 1/3], Y \sim U[-2/3, 2/3]$.}
  }
  {\includegraphics[width=0.3\linewidth]{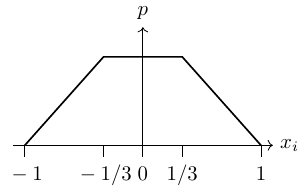}}
\end{figure}

\FloatBarrier
\newpage
\section{Qualitative Results}\label{sec:qualitativeresults}
\FloatBarrier
\begin{figure}[htbp]
\floatconts
  {fig:qualitativeoutputs}
  {\caption{We display an axial slice of three volumes. The first column shows T1ce-sequence and the ground truth segmentation. Then we display three outputs E1-E3 of the ensemble for each of the models and finally the mean- and (normalized) variance 
  map across the ensemble of size 15.}%
  }
  {
\newcommand{\rb}[1]{\rotatebox[origin=l]{90}{\scriptsize #1}}
\newcommand{\mysize}{0.103\linewidth}
\begin{tabular}{l|lllll|l}%
T1ce/GT 
& E1 & E2 & E3 & Mean & Variance 
\\
   \includegraphics[width=\mysize]{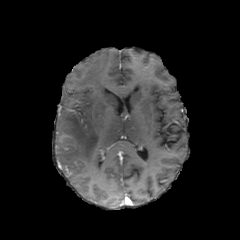} 
&  \includegraphics[width=\mysize]{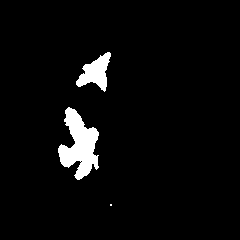} 
&  \includegraphics[width=\mysize]{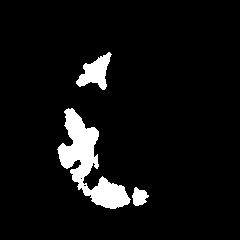} 
&  \includegraphics[width=\mysize]{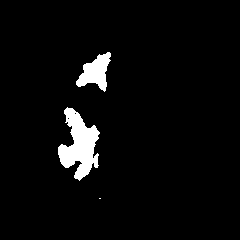} 
&  \includegraphics[width=\mysize]{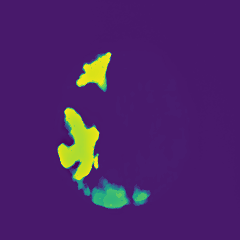} 
&  \includegraphics[width=\mysize]{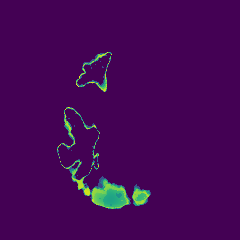} 
& \rb{PatchDDM}

\\
   \includegraphics[width=\mysize]{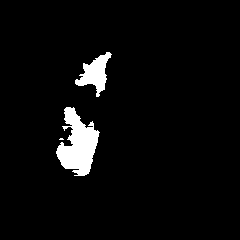} 
&  \includegraphics[width=\mysize]{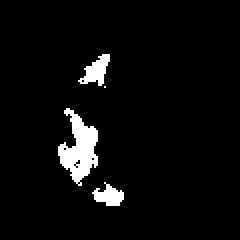} 
&  \includegraphics[width=\mysize]{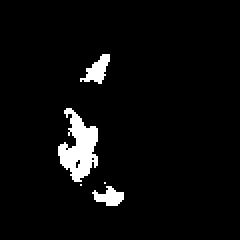} 
&  \includegraphics[width=\mysize]{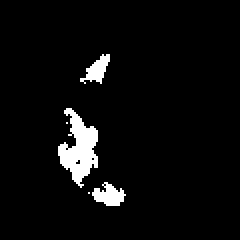} 
&  \includegraphics[width=\mysize]{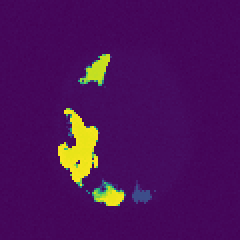} 
&  \includegraphics[width=\mysize]{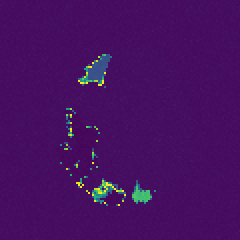} 
& \rb{HalfRes}

\\
&  \includegraphics[width=\mysize]{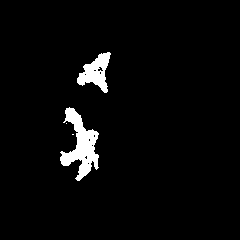} 
&  \includegraphics[width=\mysize]{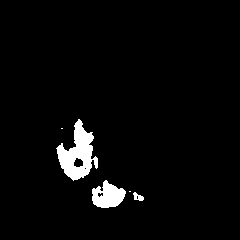} 
&  \includegraphics[width=\mysize]{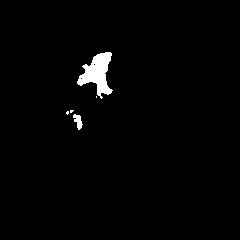} 
&  \includegraphics[width=\mysize]{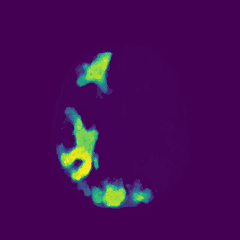} 
& \includegraphics[width=\mysize]{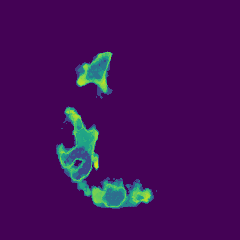} 
& \rb{FullRes}

\\
\\ %
\includegraphics[width=\mysize]{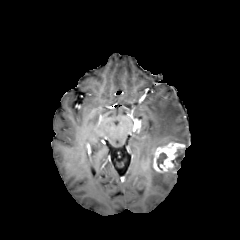} 
&  \includegraphics[width=\mysize]{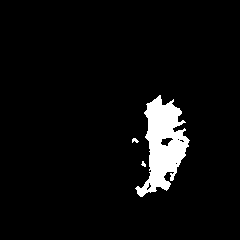} 
&  \includegraphics[width=\mysize]{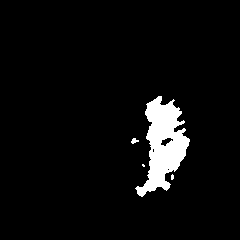} 
&  \includegraphics[width=\mysize]{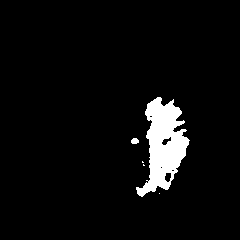} 
&  \includegraphics[width=\mysize]{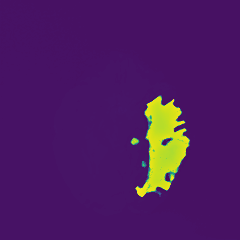} 
&  \includegraphics[width=\mysize]{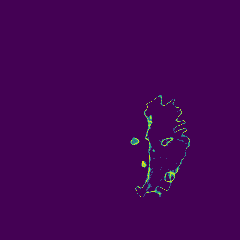} 
& \rb{PatchDDM}

\\
   \includegraphics[width=\mysize]{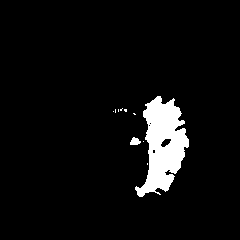} 
&  \includegraphics[width=\mysize]{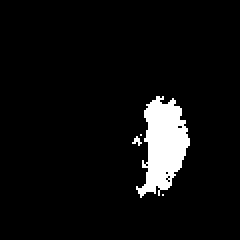} 
&  \includegraphics[width=\mysize]{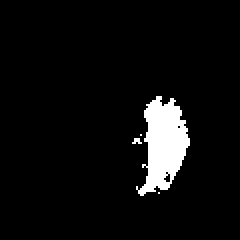} 
&  \includegraphics[width=\mysize]{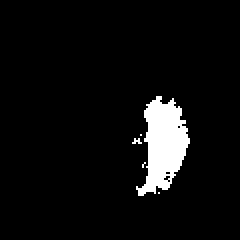} 
&  \includegraphics[width=\mysize]{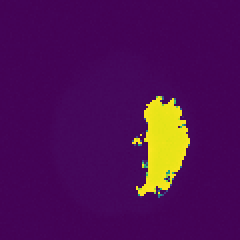} 
& \includegraphics[width=\mysize]{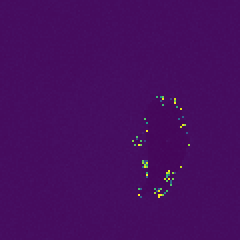} 
& \rb{HalfRes}

\\
&  \includegraphics[width=\mysize]{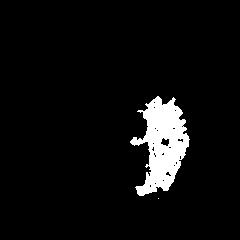} 
&  \includegraphics[width=\mysize]{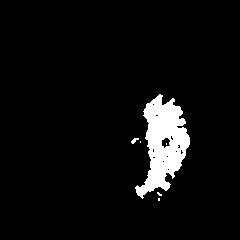} 
&  \includegraphics[width=\mysize]{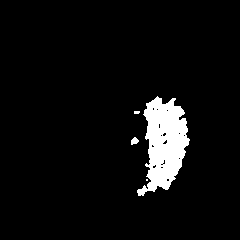} 
&  \includegraphics[width=\mysize]{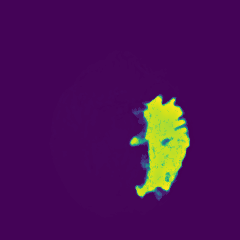} 
& \includegraphics[width=\mysize]{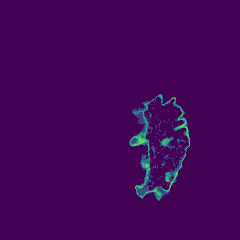} 
& \rb{FullRes}

\\
\\
\includegraphics[width=\mysize]{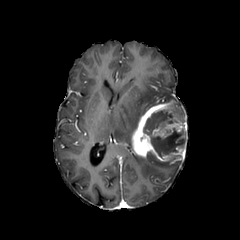} 
&  \includegraphics[width=\mysize]{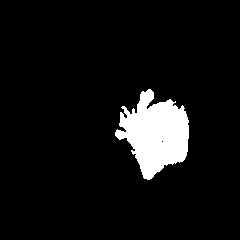} 
&  \includegraphics[width=\mysize]{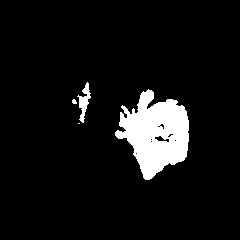} 
&  \includegraphics[width=\mysize]{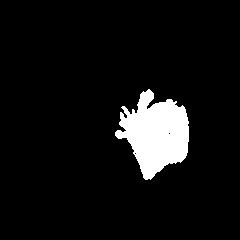} 
&  \includegraphics[width=\mysize]{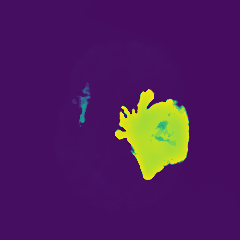} 
&  \includegraphics[width=\mysize]{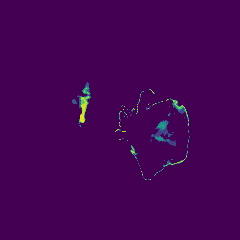} 
& \rb{PatchDDM}

\\
   \includegraphics[width=\mysize]{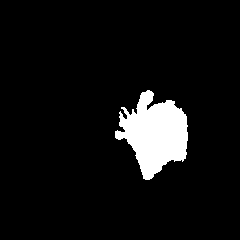} 
&  \includegraphics[width=\mysize]{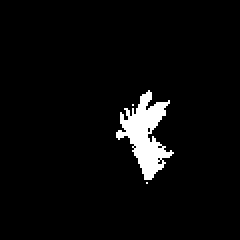} 
&  \includegraphics[width=\mysize]{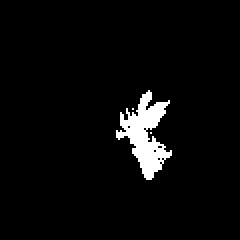} 
&  \includegraphics[width=\mysize]{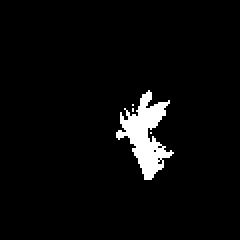} 
&  \includegraphics[width=\mysize]{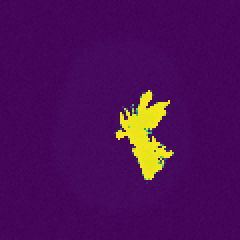} 
&  \includegraphics[width=\mysize]{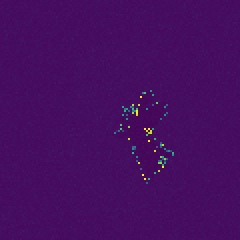} 
& \rb{HalfRes}

\\
&  \includegraphics[width=\mysize]{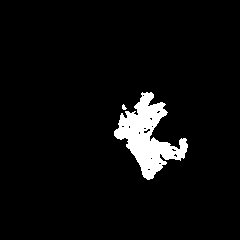} 
&  \includegraphics[width=\mysize]{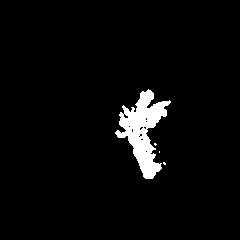} 
&  \includegraphics[width=\mysize]{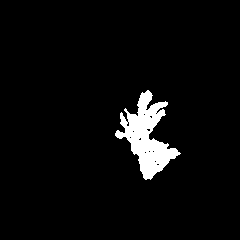} 
&  \includegraphics[width=\mysize]{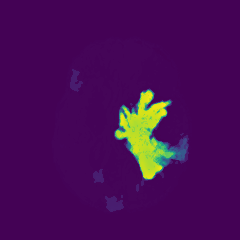} 
& \includegraphics[width=\mysize]{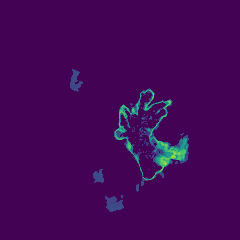} 
& \rb{FullRes}
\end{tabular}
  }
\end{figure}

\FloatBarrier
\newpage
\section{Training Progress}\label{sec:trainingprogress}
\FloatBarrier

\begin{figure}[htbp]
\floatconts
  {fig:patch_training}
  {\caption{Performance of our method \emph{PatchDDM} on the validation- and test set over the course of the training. The $x$-axis indicates the number of training iterations as a multiple of 1000.}}
  {\includegraphics[width=0.7\linewidth]{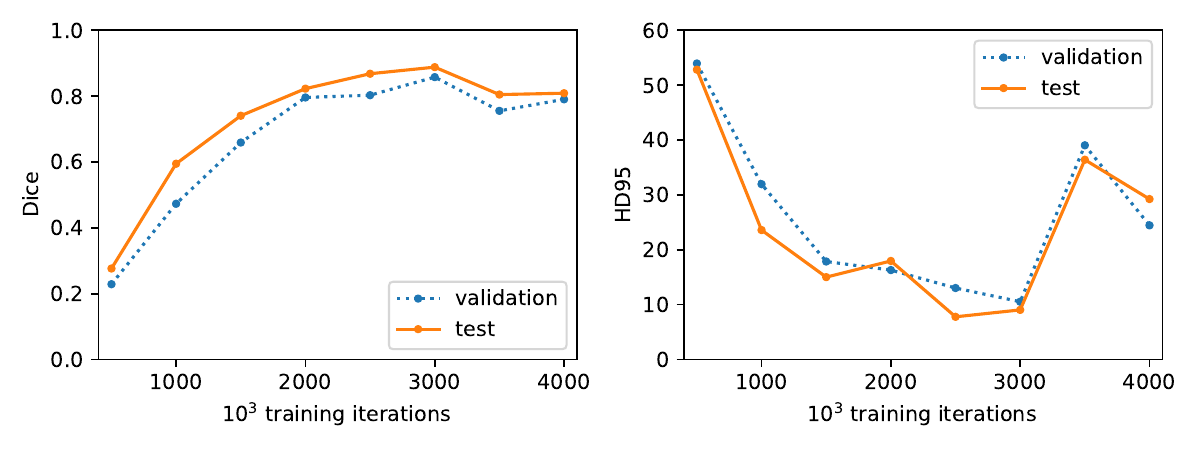}}
\end{figure}

\FloatBarrier
\section{Single Evaluation Scores}\label{sec:singleevaluationscores}
\FloatBarrier

\begin{table}[htb]
 \floatconts
 {tab:singlescores}%
 {\caption{Segmentation scores of our methods and nnU-Net on different metrics on our test set based on a single evaluation.
 }}%
 {\begin{tabular}{lll}%
 \bfseries Method & \bfseries Dice  & \bfseries HD95 %
 \\ \hline
 \emph{FullRes}    & $    0.82\pm 0.12 $ & $               16.80\pm            18.96$ %
 \\
 \emph{HalfRes}    & $    0.86\pm 0.09 $ & $\hphantom{0}{6.61}\pm \hphantom{0}{9.37}$ %
 \\
 \emph{PatchDDM} & $  {0.88\pm 0.07}$ & $    \hphantom{0}9.04\pm \hphantom{0}8.75$ %
 \\ \hline
 nnU-Net     & $    0.96\pm 0.02 $ & $    \hphantom{0}1.24\pm \hphantom{0}0.48$ %
 \\
 \end{tabular}}
 \end{table}

\FloatBarrier

\newpage
\section{Ensembling Scores}\label{sec:ensemblingscores}
\FloatBarrier

\begin{table}[htbp]
\floatconts
{tab:ensemblingscores}%
{\caption{Segmentation scores of the three methods with various ensemble sizes.}}%
{\begin{tabular}{ll|lllll}
Method &  Ensemble size  &    1     & 3       & 5       & 7       & 15     \\ \hline
\emph{FullRes} 
& Dice       & $ 0.821$ & $0.846$ & $0.849$ & $0.851$ & $0.856$\\
& HD95       & $16.80 $ & $9.91 $ & $8.00 $ & $7.72 $ & $8.13 $\\ \hline
\emph{HalfRes} 
& Dice       &  $0.862$ & $0.858$ & $0.860$ & $0.859$ & $0.862$\\
& HD95       &  $6.61 $ & $7.00 $ & $6.79 $ & $6.70 $ & $6.65 $\\ \hline
\emph{PatchDDM} 
& Dice       &  $0.888$ & $0.894$ & $0.892$ & $0.897$ & $0.899$\\
& HD95       &  $9.04 $ & $9.94 $ & $8.49 $ & $7.67 $ & $7.34 $\\ %
\end{tabular}}
\end{table}

\end{document}